\documentclass[pdflatex,sn-nature]{sn-jnl}

\usepackage{graphicx}%
\usepackage{multirow}%
\usepackage{amsmath,amssymb,amsfonts}%
\usepackage{amsthm}%
\usepackage{mathrsfs}%
\usepackage[title]{appendix}%
\usepackage{xcolor}%
\usepackage{textcomp}%
\usepackage{manyfoot}%
\usepackage{booktabs}%
\usepackage{algorithm}%
\usepackage{algorithmicx}%
\usepackage{algpseudocode}%
\usepackage{listings}%
\usepackage{main-defs}
\usepackage{subcaption}

\theoremstyle{thmstyleone}%
\theoremstyle{thmstyletwo}%

\theoremstyle{thmstylethree}%

\begin{document}

\title[A Lightweight Foundation Model for Collider Physics with Multi-Domain Adaptation]{A Lightweight Foundation Model for Collider Physics with Multi-Domain Adaptation}


\author*[1]{\fnm{Liangyu} \sur{Wu}}\email{liangyu.wu@stanford.edu}

\author*[2]{\fnm{Qibin} \sur{Liu}}\email{qibin@slac.stanford.edu}

\author[1]{\fnm{Alexander} \sur{Yue}}\email{alexyue@stanford.edu}

\author[2]{\fnm{Julia} \sur{Gonski}}\email{jgonski@slac.stanford.edu}

\affil[1]{\orgdiv{Department of Physics}, \orgname{Stanford University}, \orgaddress{\street{450 Jane Stanford Way}, \city{Stanford}, \postcode{94305}, \state{CA}, \country{USA}}}

\affil[2]{\orgname{SLAC National Accelerator Laboratory}, \orgaddress{\street{2575 Sand Hill Road}, \city{Menlo Park}, \postcode{94025}, \state{CA}, \country{USA}}}


\abstract{We present a lightweight approach to foundation modeling (\textbf{NEXUS}) that leverages pre-trained learning from collider physics data towards out-of-domain tasks in other scientific datasets, using a fully connected autoencoder model with approximately 3 million parameters.
The model pre-trains with no supervision over a large-scale collision dataset from the Large Hadron Collider modeled by charged particle track features. 
Downstream tasks for collider analyses, such as kinematic regression and event classification, are developed on pre-trained model weights and achieve improved accuracy with only small labeled datasets when compared to equivalent architectures trained from scratch.
The benefits of pre-training are additionally investigated through latent space interpretation and application to other domains, including gravitational waves, flood forecasting, and neural activity.
Furthermore, the relative computational simplicity of NEXUS is demonstrated compared to transformer approaches at comparable scale, opening the door to power-efficient inference and real-time or edge applications of foundation models in scientific experiments.}




\maketitle

\section{Introduction}\label{sec1}

High energy colliders offer a unique window into fundamental physics, probing the limits of accessible energy scales with highly generic detectors.
However, the complexity and enormous data rates of these detectors present challenges for data acquisition, filtering, reconstruction, and analysis. 
Artificial intelligence and machine learning (AI/ML) have proven to be uniquely useful tools to exploit these experiments for scientific potential~\cite{albertsson2019machinelearninghighenergy,duarte2025machinelearning,Karagiorgi2022,Radovic2018}.

As AI/ML capabilities advance alongside available computational power, the full statistical power of collider datasets can be exploited via large-scale training of foundation models~\cite{bommasani2022opportunitiesrisksfoundationmodels,zhou2023comprehensivesurveypretrainedfoundation}. 
A backbone model pre-trained at scale (i.e., using a high fraction of available training instances and computing resources) can generate a representation of the dataset which is transferred and fine-tuned for a range of smaller-scale downstream tasks~\cite{KUCHERA2019156,Chappell_2022,Dreyer2022}, thus enabling diverse objectives to benefit from a shared learned representation.
This reduces the need to train multiple task-specific models from scratch with comparably intensive training, allowing for higher performance at a wider variety of tasks within a fixed computational budget.
Foundation models also help mitigate the challenge of limited available training instances, particularly relevant in the scientific context where processes of interest are rare and experimental data collection is finite. 

Foundation models are being actively studied in high energy physics (HEP), with applications in jet physics~\cite{golling2024maskedparticlemodelingsets,PhysRevD.111.054015,Bhimji_2026,Birk_2024}, searches for new phenomena~\cite{hsu2026evenetfoundationmodelparticle}, and more~\cite{Mikuni_2025}.
Moreover, foundation models pre-trained on HEP data have been shown to enable effective transfer learning to other scientific domains~\cite{mikuni2025omnicosmostransferringparticlephysics}. 
However, such transfer typically relies on some degree of training supervision, as well as on resource-intensive architectures such as transformers~\cite{vaswani2023attentionneed}, which are computationally expensive to train and deploy, often involving $\mathcal{O}(10)$ million parameters and datasets containing $\mathcal{O}(100)$ million training instances.
Attention-free architectures, such as those relying on multi-layer perceptrons, can achieve competitive performance on specific tasks without the overhead of self-attention~\cite{tolstikhin2021mlpmixerallmlparchitecturevision,touvron2021resmlpfeedforwardnetworksimage,liu2021payattentionmlps}, 
though this potential remains largely unexplored in the context of pre-training and transfer learning, especially in collider physics. 

This work presents an approach to retain the performance benefits of large-scale pre-training without the computational complexity of transformer-based architectures, thereby reducing the computational cost associated with both model development and deployment.
We introduce a \textbf{N}eural network-based \textbf{E}ncoder$\boldsymbol{\times}$decoder for \textbf{U}niversal \textbf{S}ensitivity \textbf{(NEXUS)}, based on an autoencoder architecture with fully connected layers. 
It is pre-trained in a fully unsupervised manner on collider events modeled by low-level features, specifically observables from charged-particle tracks. 
With 3 million parameters and 20 million training instances, NEXUS demonstrates transfer learning across a range of collider physics tasks and extends to other scientific domains. 
These results indicate that, for certain modalities, large-scale pre-training with comparatively simple deep neural network (DNN) architectures can still yield substantial improvements on downstream transfer tasks. 
This challenges the prevailing view that foundation models necessarily require computationally intensive components such as attention mechanisms, and suggests a path toward lower-power and lower-cost foundation models for scientific applications, including deployment in real-time and other resource-constrained environments.

\section{Results}
\label{sec:results}

\begin{figure}[h!]
\begin{center}
\includegraphics[width=1.0\textwidth]{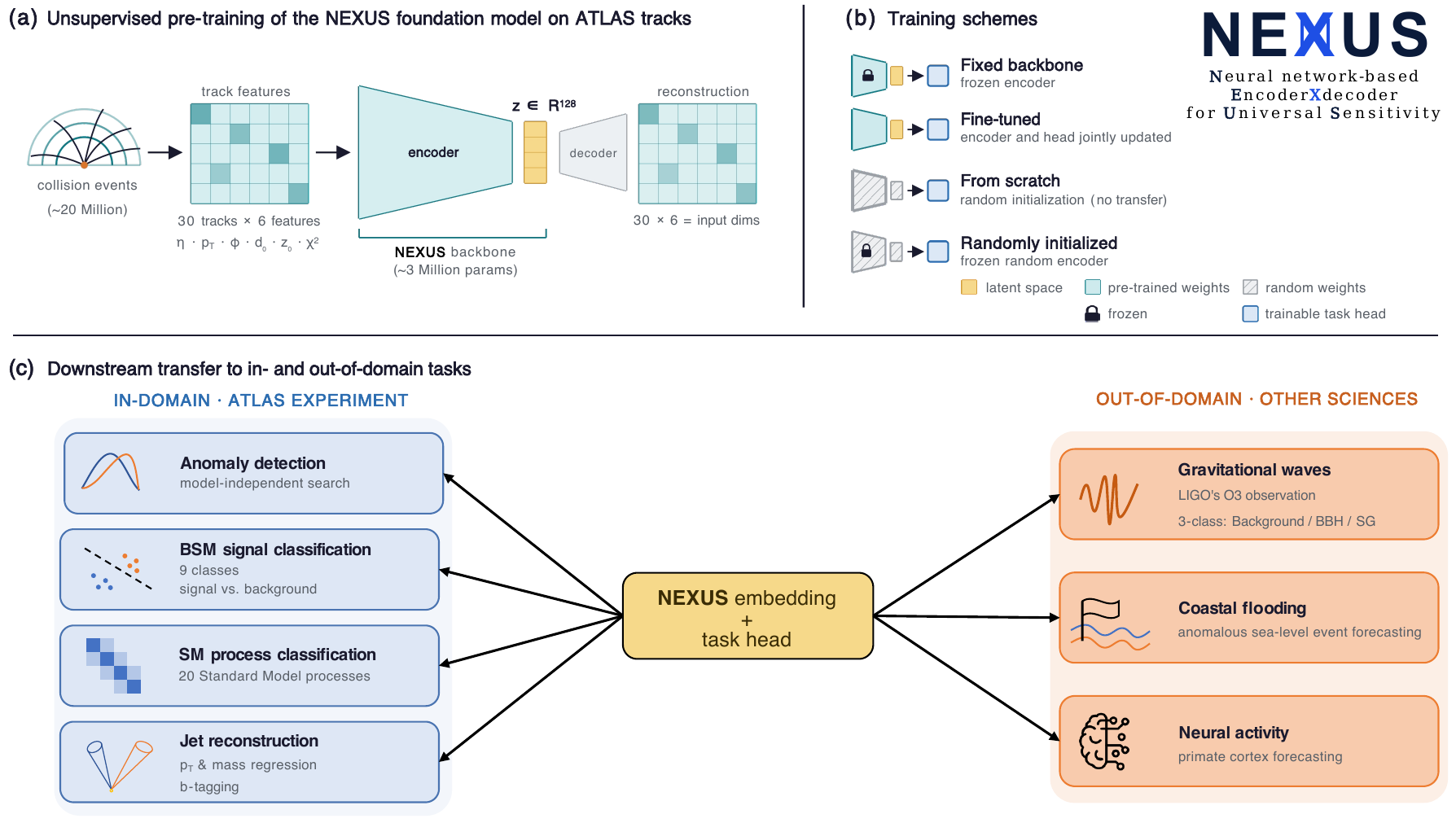}
\end{center}
\caption{\textbf{Overview of the NEXUS foundation model and its transfer to in- and
  out-of-domain downstream tasks.}
  \textbf{(a)} Unsupervised pre-training over ATLAS collision events represented by 
  inner-detector tracks as a $30\times6$ tensor (up to 30 leading-\pt~tracks, each
  described by six features: $\eta$, \pt, $\phi$, $d_0$, $z_0$ and $\chi^2$). 
  \textbf{(b)} Diagram of the NEXUS backbone usage in the four training schemes for downstream studies, namely fixed backbone, fine-tuned, from scratch, and randomly initialized. 
  Teal denotes pre-trained weights, hatching denotes random weights, a padlock denotes frozen weights, and the blue square denotes the trainable task head.
  \textbf{(c)} Downstream transfer. The pre-trained latent feeds a
  lightweight task head for both in-domain ATLAS (blue) and out-of-domain (orange) tasks.
  }\label{fig:nexus_overview}
\end{figure}

A foundation model is characterized by a pre-trained model that learns broadly transferable representations that can be adapted to a wide range of downstream tasks with minimal additional training.
Figure~\ref{fig:nexus_overview} provides a visual representation of the NEXUS foundation model and its subsequent downstream applications.
The backbone model comprises the encoder and its embedding space, and the downstream model comprises the backbone plus a linear probe adjusted to match the output dimension of each downstream task. 
To fully explore the transfer learning space of the model, the pre-trained backbone is transferred to four tasks in collider physics and three out-of-domain tasks.

The benefit of pre-training is quantified by measuring the extent to which a pre-trained model performs better at a given downstream task with a low number of training instances as compared to a model trained from scratch. 
Testing this hypothesis requires the examination of three different model training schemes:
\textit{fixed backbone}, where the pre-trained weights are fixed and only the linear probe weights are adjusted during training; \textit{fine-tuned}, where the full downstream model can be updated during training starting from the pre-training weight values; and \textit{from scratch}, where the downstream model is trained for a given task starting from random weights.

\subsection{Collider Physics}

Charged particle tracks provide a useful description of proton–proton collisions at the ATLAS experiment~\cite{TheATLASCollaboration_2008} at the Large Hadron Collider (LHC), capturing the momenta and trajectories of charged-particles with excellent spatial resolution. 
Although tracks do not fully characterize the final state of a collision, they are used here as a single modality to test DNN-based foundation models using set-based inputs.
The application of NEXUS focuses on key tasks in collider physics experimental research, specifically the reconstruction of raw detector signals into particles and the selection of interesting events based on their particle content. 

\subsubsection{Anomaly Detection}

Anomaly detection, or the recognition of unusual elements in a dataset without signal-specific priors, is an essential component of the search program at colliders as the nature of beyond the Standard Model (BSM) physics is unknown~\cite{BELIS2024100091}.
Autoencoders are commonly used for this task as their unsupervised training creates a latent space where data-like events are clustered and unusual elements will fall out of distribution. 
As a first opportunity to understand the NEXUS capacity for transfer learning, its learned clustering of ATLAS events in the backbone latent space is studied using anomaly detection metrics. 

\begin{figure}[h!]
\begin{center}
\includegraphics[width=0.6\textwidth]{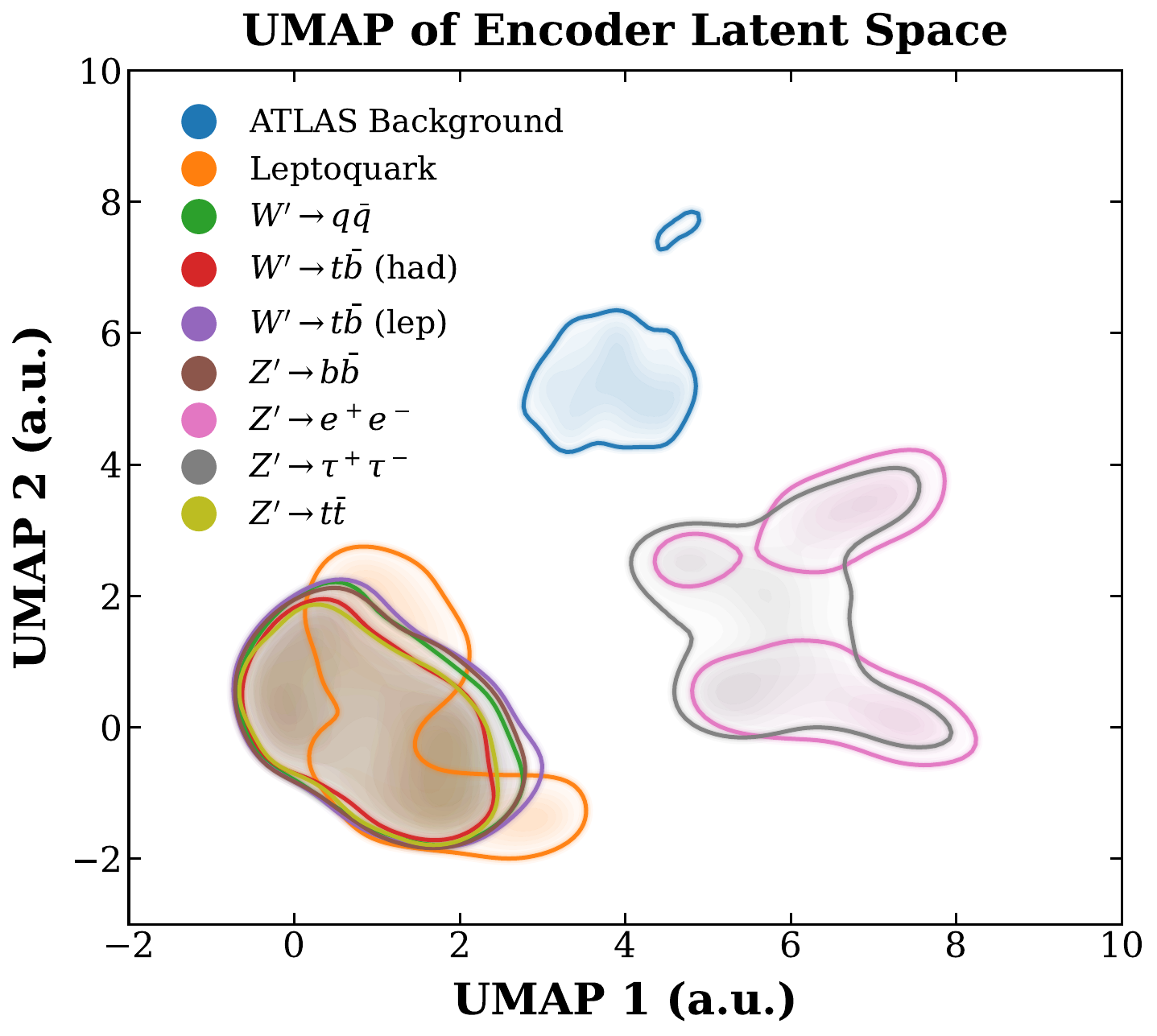}
\end{center}
\caption{\textbf{Encoder latent space of the nine event classes.} Two-dimensional UMAP projection of the frozen NEXUS backbone latents: the Standard Model background (blue) is isolated, the leptonic $Z'\to ee/\tau\tau$ signals form a distinct region, and the jet-rich hadronic signals overlap at the lower left. Shaded regions show the per-class density. Contours enclose the densest 68\% of each class.}
\label{fig:perf_classification}
\end{figure}

A two-dimensional Uniform Manifold Approximation and Projection (UMAP)~\cite{Healy2024} projection is used to visualize the latent space structure for ATLAS collision data and eight BSM signals from Monte Carlo (MC) simulation (Figure~\ref{fig:perf_classification}). 
Three broad regions emerge in this label-agnostic projection: the Standard Model (SM) background (represented by data), the two leptonically decaying resonances ($Z'\to e^{+}e^{-}$ and $Z'\to\tau^{+}\tau^{-}$), and the topologically similar jet-rich signals (leptoquark, $W'\to q\bar{q}$, $W'\to tb$ with hadronic and leptonic top decays, $Z'\to b\bar{b}$ and $Z'\to t\bar{t}$).
This organization reveals the basic topological features learned by NEXUS: the background and the leptonic signals are resolved with ease, whereas the residual confusion is concentrated among the topologically similar hadronic final states with similar track signatures.
The negative log-density of a kernel-density estimate (KDE) fit to the 2D UMAP projection, where events in low-density, background-sparse regions receive higher anomaly scores, provides an area-under-curve (AUC) greater than 0.95 across all signals, indicating useful anomaly classification in the NEXUS latent space. 

\subsubsection{Event Classification}

Two supervised classification tasks for NEXUS are considered: 
a nine-way supervised classification of BSM signals from the background, and an in-distribution task to classify 20 SM processes spanning the dominant LHC backgrounds (Section~\ref{sec:methods}).
Foundation model performance is summarized using training dataset scan plots, where the number of training instances is varied from 100\% to 0.1\% of the available training set, and the performance is evaluated across the three training configurations.
Figures~\ref{fig:FMplots_ATLAS_exotic} and~\ref{fig:FMplots_ATLAS_sm} show the results from the BSM and SM classification tasks, respectively. 
For each task, the two schemes initialized from pre-trained weights, the fixed backbone and the fine-tuned model, both outperform the from-scratch baseline at low numbers of training instances, revealing the benefit of pre-training. 
In the extreme low-data regime, the fixed backbone is typically strongest, since its small number of trainable parameters resists overfitting~\cite{kumar2022finetuningdistortpretrainedfeatures}.
When using the full training dataset, the from-scratch model generally approaches or exceeds the fine-tuned performance, indicating that with sufficient data the architecture converges to its optimal performance regardless of initialization.

\subsubsection{Jet Reconstruction}

\begin{figure}[htbp]
\centering

\begin{subfigure}[t]{0.48\textwidth}
  \centering
  \includegraphics[width=\textwidth]{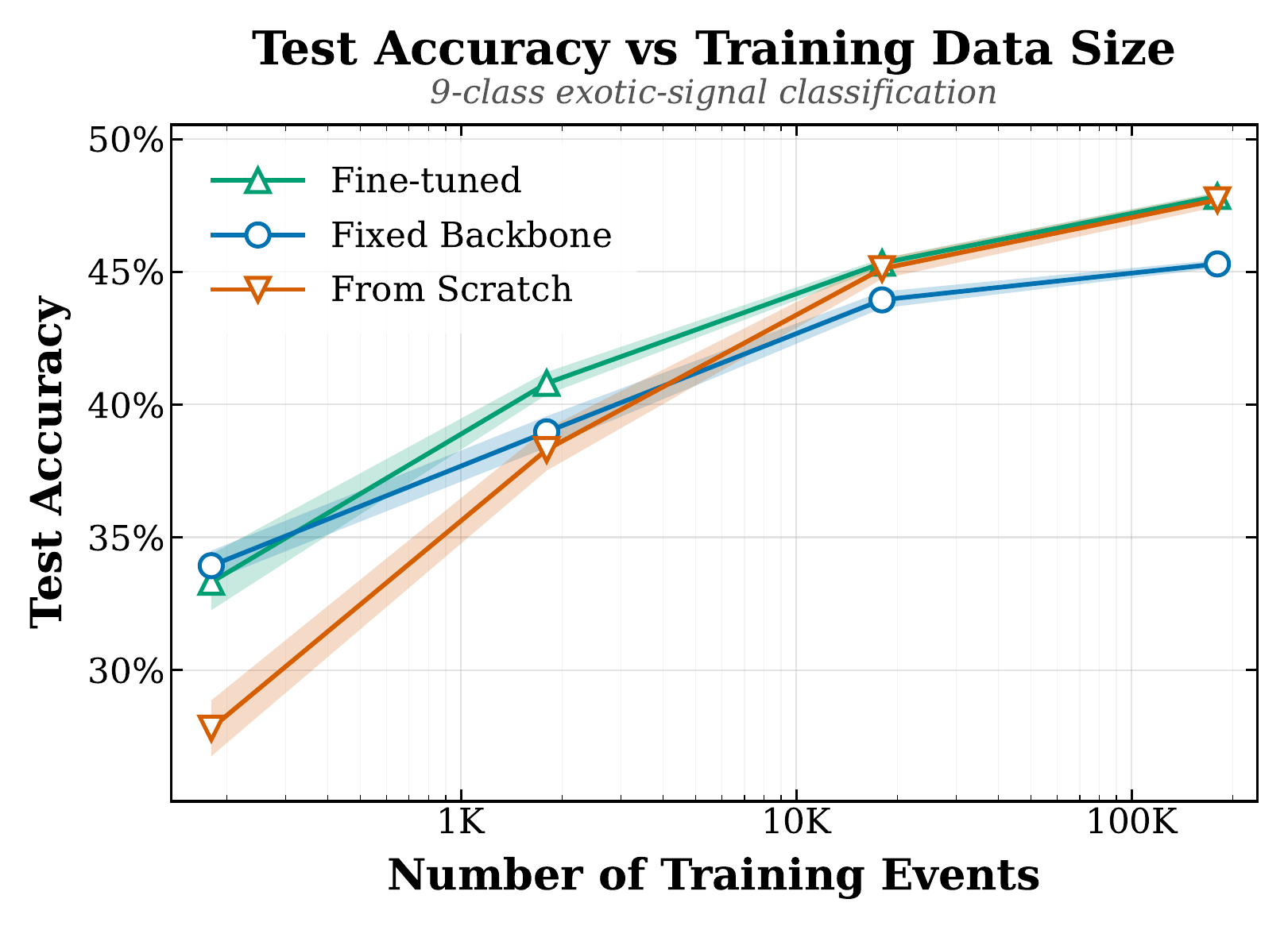}
  \caption{BSM classification.}
  \label{fig:FMplots_ATLAS_exotic}
\end{subfigure}\hfill
\begin{subfigure}[t]{0.48\textwidth}
  \centering
  \includegraphics[width=\textwidth]{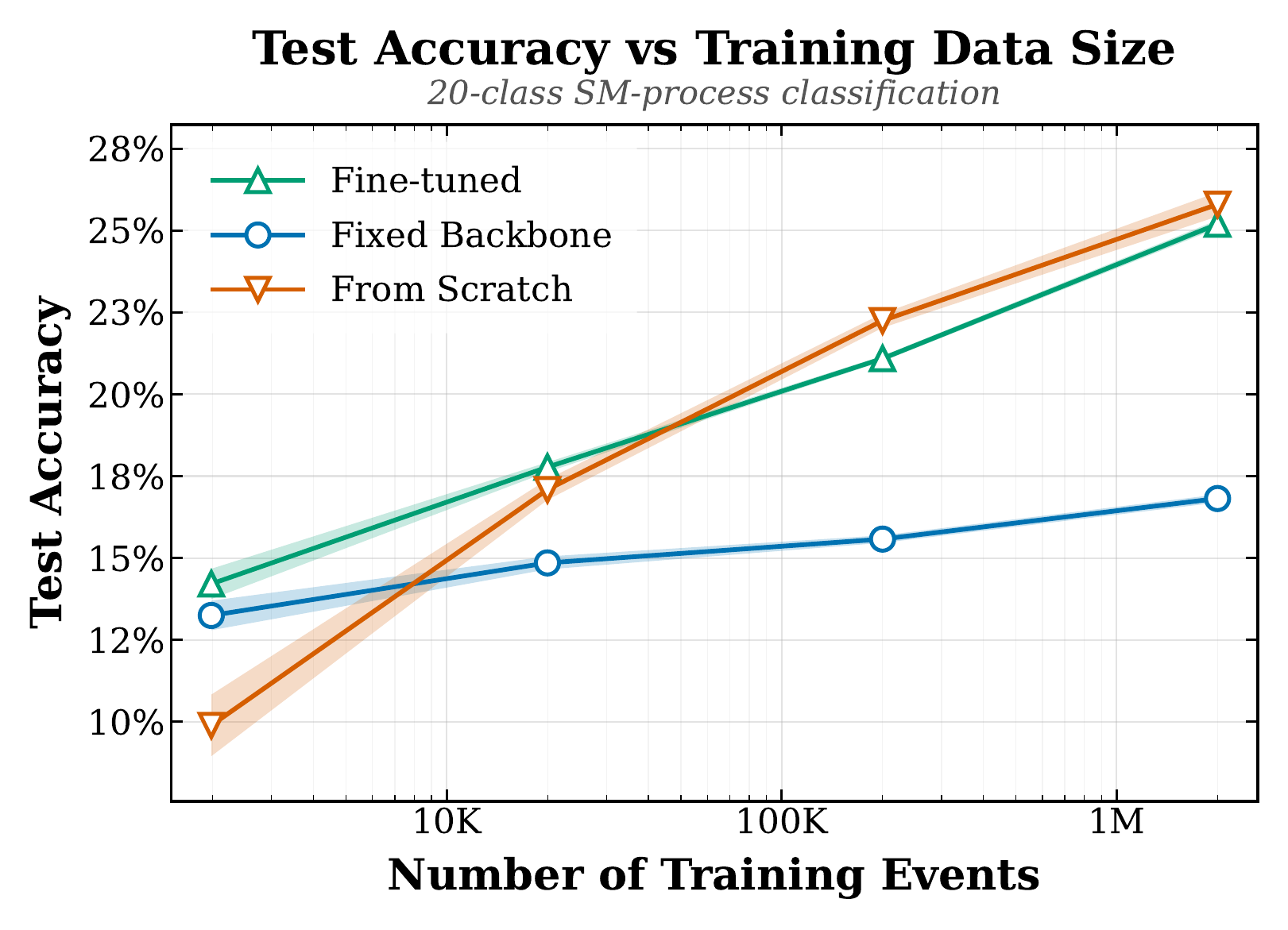}
  \caption{SM classification.}
  \label{fig:FMplots_ATLAS_sm}
\end{subfigure}


\begin{subfigure}[t]{0.48\textwidth}
  \centering
  \includegraphics[width=\textwidth]{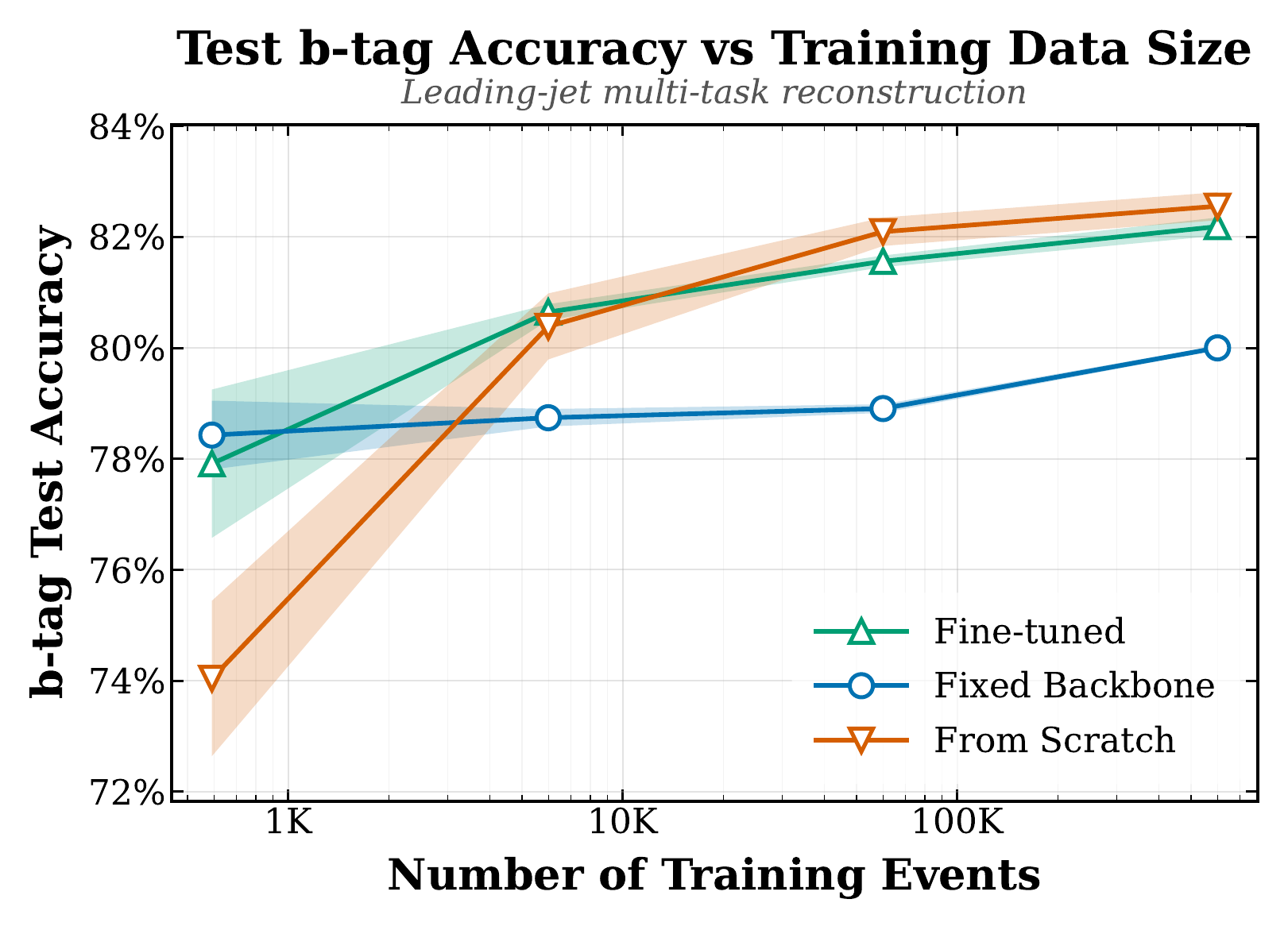}
  \caption{Jet $b$-tag accuracy.}
  \label{fig:FMplots_ATLAS_btag}
\end{subfigure}\hfill
\begin{subfigure}[t]{0.48\textwidth}
  \centering
  \includegraphics[width=\textwidth]{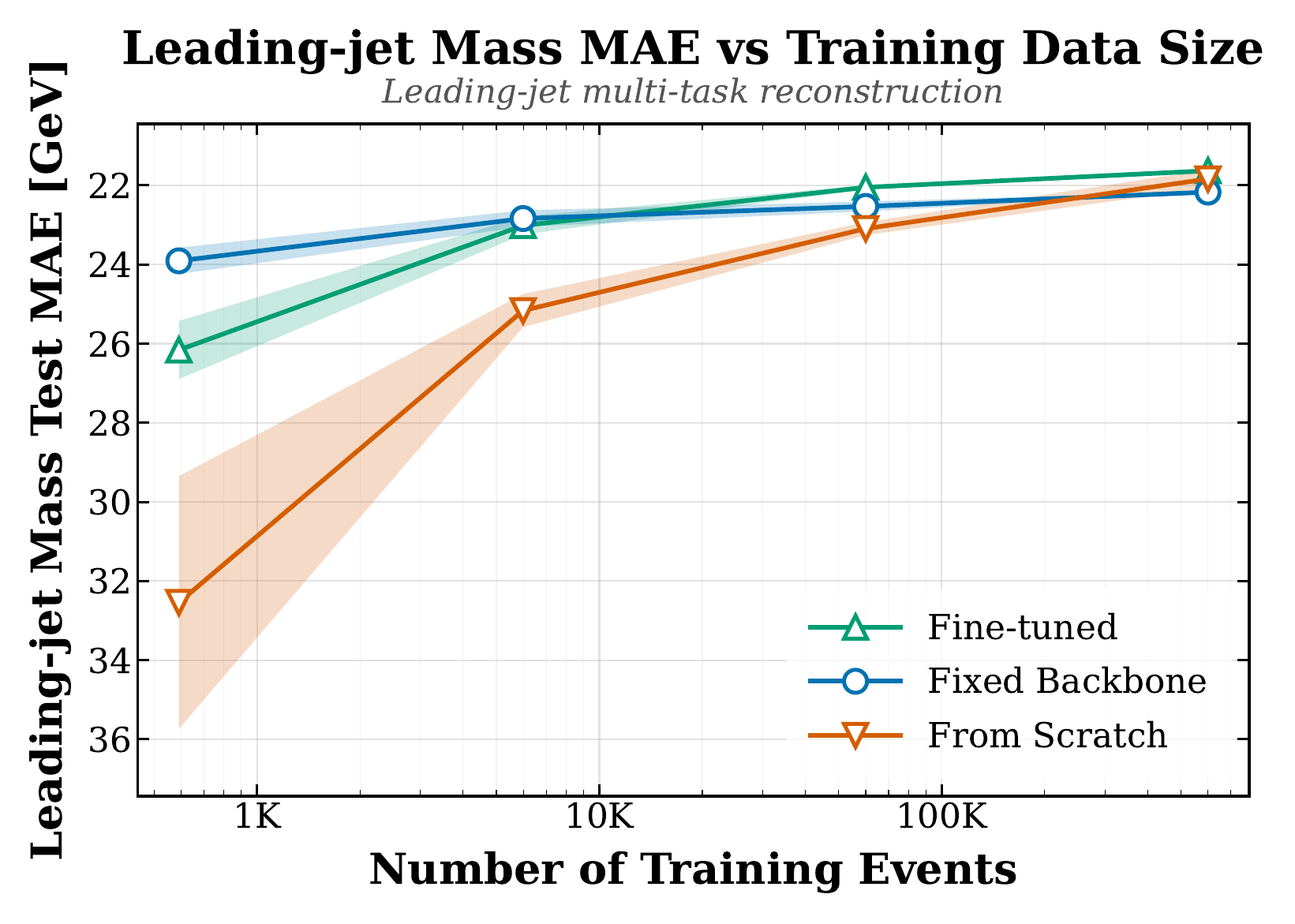}
  \caption{Jet mass regression.}
  \label{fig:FMplots_ATLAS_mass}
\end{subfigure}


\begin{subfigure}[t]{0.48\textwidth}
  \centering
  \includegraphics[width=\textwidth]{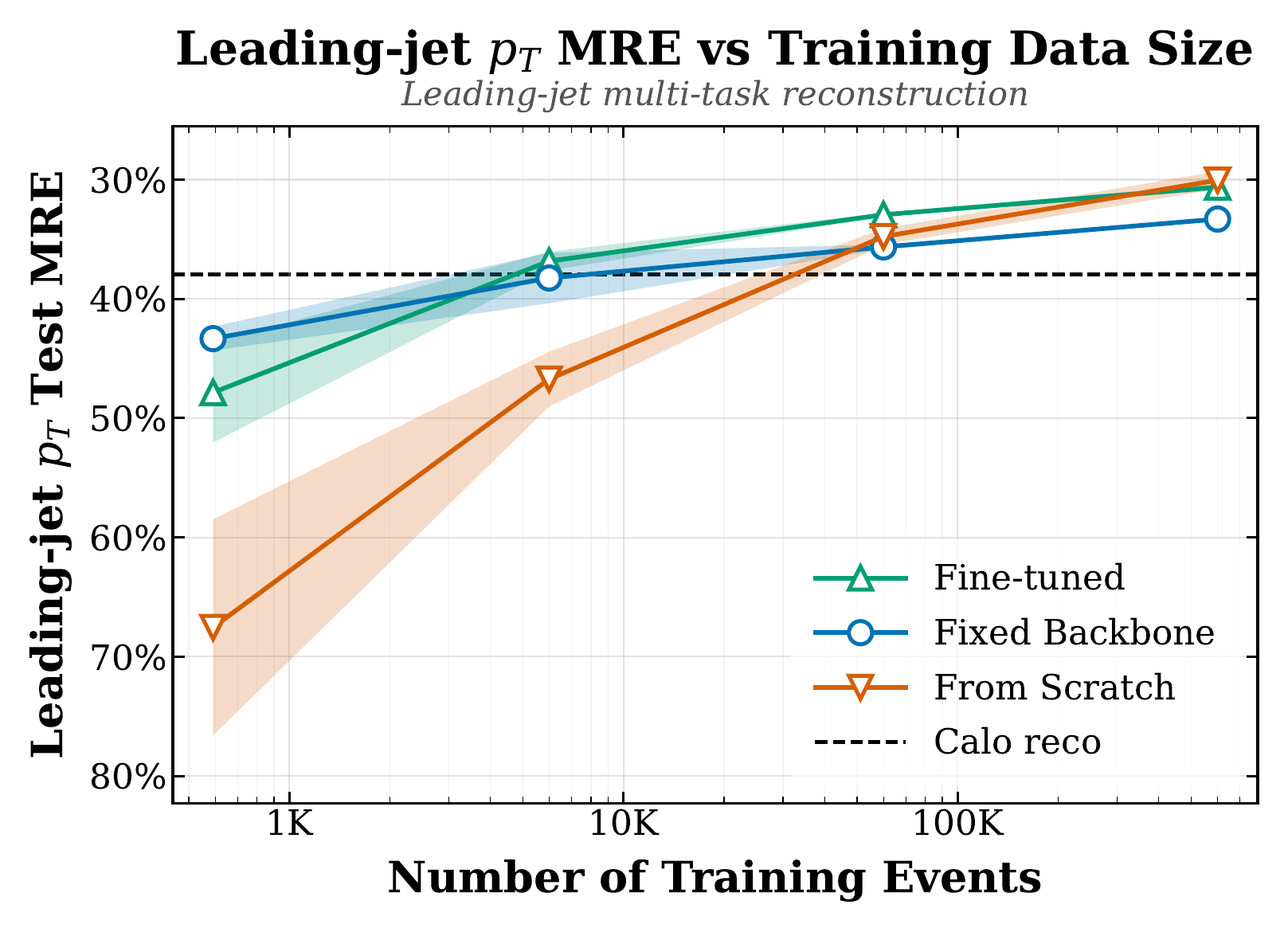}
  \caption{Jet \pt~regression.}
  \label{fig:FMplots_ATLAS_pt}
\end{subfigure}

\caption{\textbf{Foundation model scaling on the downstream collider tasks.} For each task the three training schemes (\textit{fixed backbone}, \textit{fine-tuned}, and \textit{from scratch}) are compared on identical training/validation/test splits as a function of the number of labeled training events, scanned from 100\% down to 0.1\% of the available training set. Performance is the task-appropriate test metric evaluated on the held-out test set. For the calorimeter-derived jet \pt, the resolution of the standard calorimeter-based reconstruction (``Calo reco'') is overlaid as a non-ML reference. Points and bands show the mean and standard deviation over ten random seeds.}
\label{fig:FMplots_ATLAS}
\end{figure}

A jet is a spray of particles produced by an initial quark or gluon which hadronizes in the detector, leaving charged-particle tracks as well as energy deposits in the calorimeters. 
The NEXUS jet task relates to reconstruction, namely the simultaneous regression of the jet transverse momentum \pt~and invariant mass $m$, together with identification of the presence of a $b$-quark (``$b$-tagging''). 
Because the jet \pt~is primarily a calorimeter-derived quantity (as reconstructed in the current ATLAS experiment~\cite{TheATLASCollaboration_2008}) that is inaccessible to the track-only NEXUS inputs, the resolution of the standard calorimeter-based jet reconstruction as obtained from ATLAS Open Data is reported as a non-ML reference.
Figures~\ref{fig:FMplots_ATLAS_btag},~\ref{fig:FMplots_ATLAS_mass}, and~\ref{fig:FMplots_ATLAS_pt} show the foundation model plots for the jet $b$-tagging, mass, and \pt~regression, respectively. 
The same broad conclusions about transfer learning extend to the jet reconstruction tasks; across training dataset sizes, pre-trained initialization outperforms training from scratch (especially with few labels), with fine-tuning best overall and a fixed backbone still competitive in the low-statistics regime.

\subsection{Out-of-Domain}

The NEXUS pre-trained backbone is further tested on three distinct non-collider scientific datasets, obtained from community challenges focusing on scientific anomaly detection~\cite{nsfhdr_ad} and out-of-distribution modeling~\cite{nsfhdr_mood}. 
All out-of-domain datasets consist of one-dimensional time-series values, requiring the collider track input features to be mapped to features that describe 1D waveforms. 
Table~\ref{tab:mapping} provides the mappings used in this study, which are designed to capture the same underlying detector response, while respecting analogous symmetries and information content in different coordinate systems. 
The collider variables encode the objects' geometry and kinematics in detector coordinates (e.g., angular coordinates $\phi$ and $\eta$, energy scale \pt, track displacement from the beamline described by $z_0$ and $d_0$, and track fit quality $\chi^2$). 
The waveform features provide an appropriate mapping with similar symmetry characteristics: 
locations in the time axis play the role of phase or rotation degrees of freedom (e.g., 
the time centroid for the gravitational-wave and flooding waveforms, the weighted trough location for the neural traces); standard deviation (std) and mean set an overall scale analogous to \pt; 
and shape measures (tail asymmetry, kurtosis, pulse depth) capture skew and deviations, akin to displacement and fit quality. 
The $z_0$ variable carries a task-specific displacement analogue: an inter-detector time delay for gravitational waves, a static station longitude for coastal flooding, and the local slope for neural traces.

\begin{table}[h]
\centering
\renewcommand{\arraystretch}{1.5}
\setlength{\tabcolsep}{22pt}
\begin{tabular}{c|cc}
\toprule
\textbf{ATLAS track} & \textbf{GW, CF} & \textbf{Neural activity} \\
\midrule
$\eta$ & Energy centroid & Weighted trough loc. \\
\pt & Std & Std \\
$\phi$ & Time centroid & Channel ID \\
$d_0$ & Tail asymmetry & Mean \\
$z_0$ & $\Delta t$ & Slope \\
$\chi^2$ & Kurtosis & Pulse depth \\
\bottomrule
\end{tabular}
\caption{Mapping of ATLAS track features to the 1D waveform features used for each out-of-domain task. Gravitational waves (GW) and coastal flooding (CF) share the same feature scheme, while the neural-activity task uses a distinct mapping. For GW and CF, the $z_0$ row ($\Delta t$) denotes the inter-detector time delay and the static station longitude, respectively.}
\label{tab:mapping}
\end{table}

While the out-of-domain tasks are useful to explore the boundary of transfer learning from the NEXUS model, it is not expected that the specific NEXUS architecture can provide competitive performance for widely varying tasks compared to a dedicated ML approach. For this reason, a fourth training configuration is added for the out-of-domain studies: \textit{randomly initialized}, where the NEXUS structure is used with weights that are randomly initialized and kept fixed (representing complete ignorance of the task at hand). This is useful to clearly delineate any marginal benefit provided by the collider pre-training in the absence of strong overall task performance.

\subsubsection{Gravitational Wave Anomalies}

The gravitational waves (GW) anomaly detection task is cast as a supervised three-class discrimination task over the track-mapped strain: instrumental background, modeled binary black hole (BBH) signals, and unexpected (unmodeled, burst-like) GW transients. Each strain segment, recorded by the two detectors, is summarized by the waveform descriptors of Table~\ref{tab:mapping} and arranged into the $30\times6$ track-like tensor expected by the NEXUS backbone. 

\begin{figure}[h!]
\begin{center}
\includegraphics[width=0.48\textwidth]{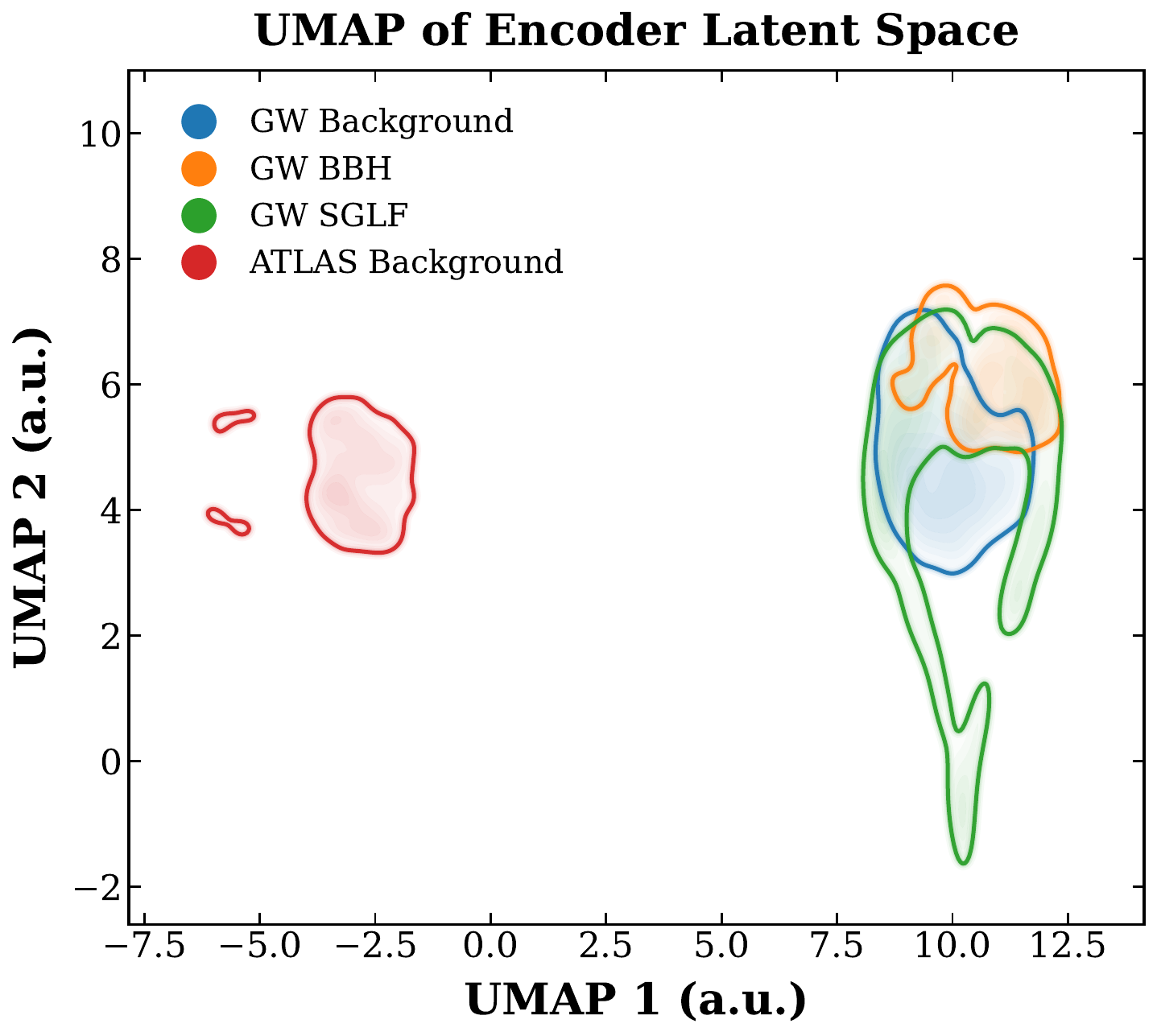}
\includegraphics[width=0.48\textwidth]{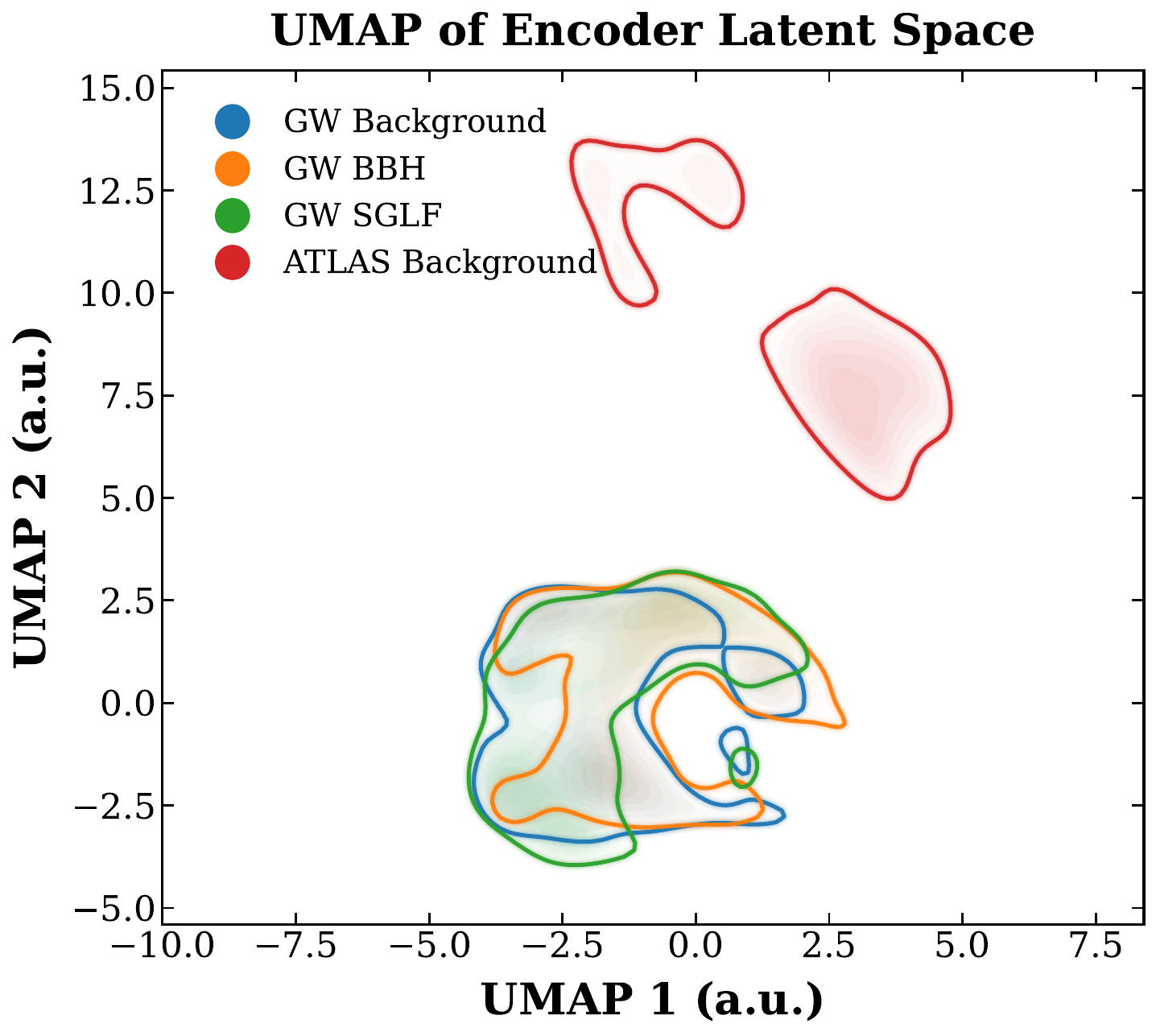}
\end{center}
\caption{\textbf{UMAP projection of collider and gravitational wave samples}, with physically motivated (left; defined in Table~\ref{tab:mapping}) vs. random (right) feature mapping. Without physically motivated feature correspondence to the track-based input of the NEXUS backbone, the latent space is unable to structure classification of the three GW classes, underlining the existence of transfer learning in the correct mapping scenario.}\label{fig:umap_gravwaves}
\end{figure}

The importance of the waveform summary features and their correspondence to the meaningfully similar NEXUS track features is assessed through a feature-permutation study, in which the assignment of GW waveform to track feature is shuffled before being passed to the encoder. Figure~\ref{fig:umap_gravwaves} shows UMAP projections of the backbone latent space of the ATLAS data and three GW classes, both under the physically motivated mapping and under a representative permuted ordering. 
Only under the physically motivated mapping do the three GW classes occupy distinct regions of the latent space, indicating that pre-training has learned a meaningful understanding of waveform structure that can be lent to out-of-domain extrapolation. 
Figure~\ref{fig:FMplots_OOD_a} shows the foundation model plot for the gravitational wave classification task, where, similarly to collider tasks, the fine-tuned model is strongest across the range, 
the from-scratch model only matches the fine-tuned one once sufficient data is available, and the randomly initialized network is weakest throughout.


\subsubsection{Coastal Flooding Prediction}

\begin{figure}[h!]
\centering
\begin{subfigure}[t]{0.48\textwidth}
\centering
\includegraphics[width=\textwidth]{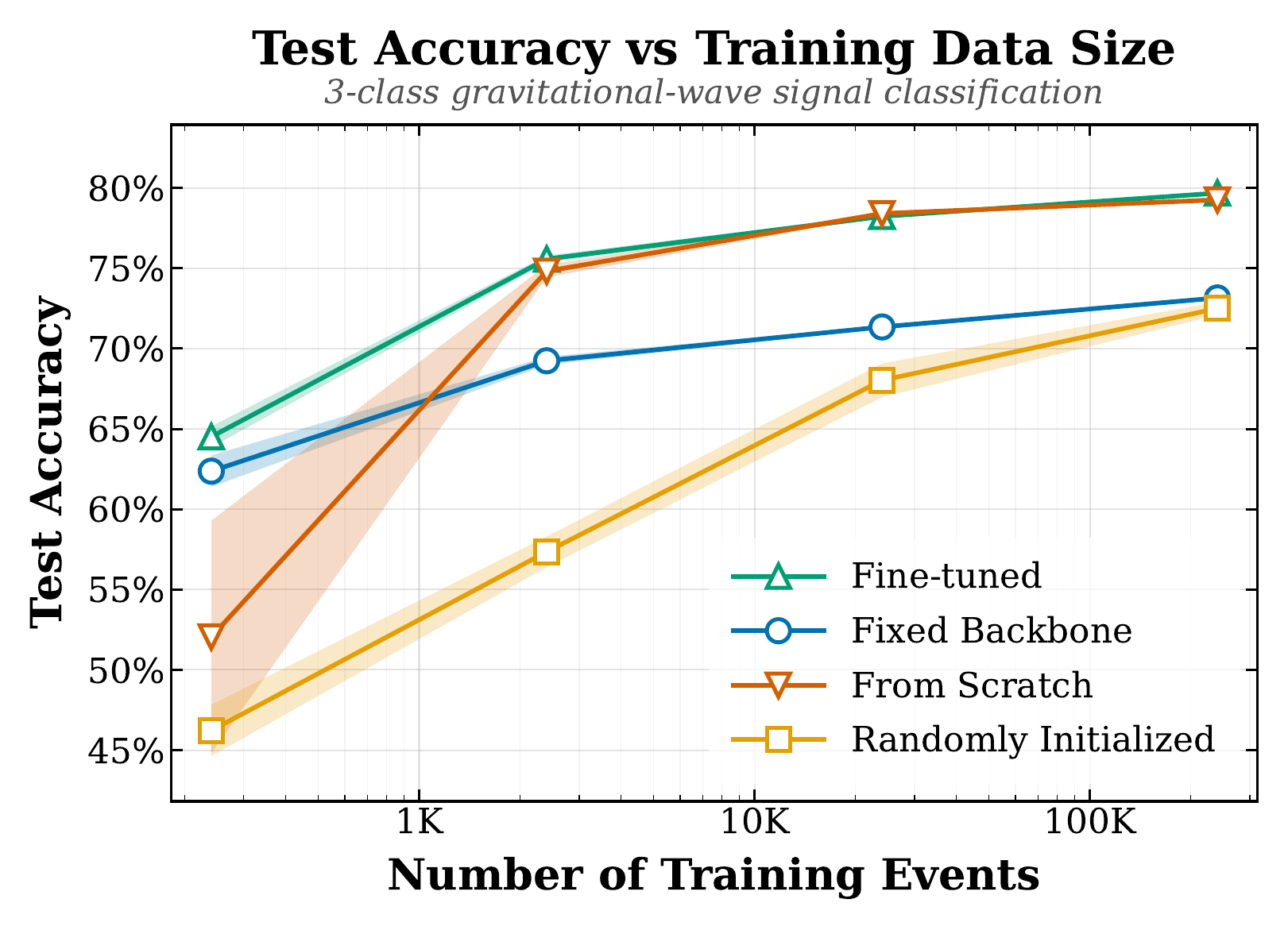}
\caption{Gravitational wave anomaly classification.}
\label{fig:FMplots_OOD_a}
\end{subfigure}\hfill
\begin{subfigure}[t]{0.48\textwidth}
\centering
\includegraphics[width=\textwidth]{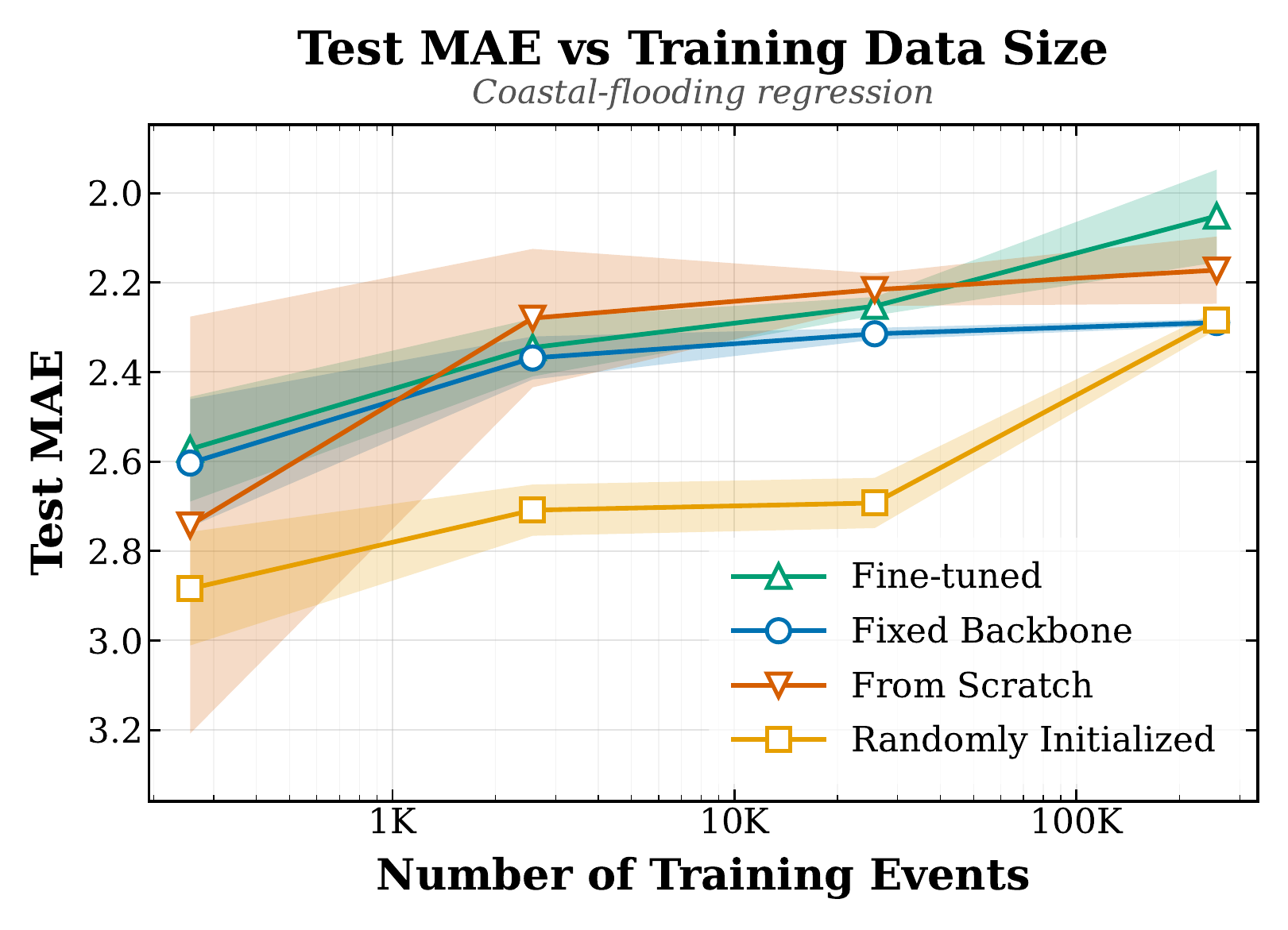}
\caption{Coastal flooding prediction.}
\label{fig:FMplots_OOD_b}
\end{subfigure}
\caption{\textbf{Foundation model scaling on the two out-of-domain tasks:} gravitational-wave anomaly
classification (left) and coastal flooding prediction (right). Training schemes and scan procedure as in Figure~\ref{fig:FMplots_ATLAS}, with the
addition of the \textit{randomly initialized} scheme. Performance
is the task-appropriate test metric evaluated on the held-out test set: three-class accuracy
for the gravitational-wave task and the mean absolute error on the predicted number of flood
days for coastal flooding. Points and bands show the mean and standard deviation over ten random seeds.}
\label{fig:FMplots_OOD}
\end{figure}

The coastal flooding task is addressed as a regression of future flood occurrence on the feature-mapped sea-level signal.
Its foundation model scaling plot is provided in Figure~\ref{fig:FMplots_OOD_b}.
A similar pattern to the gravitational wave task is observed: the pre-trained models give the lowest mean absolute error at the smallest training fraction and the fine-tuned model attains the lowest error overall, while the from-scratch and fine-tuned models become comparable at intermediate fractions. 
As anticipated for cross-domain transfer, the absolute performance for both tasks is modest and well below that of dedicated domain-specific methods, but nonetheless the pre-trained weights provide an advantage over an architecture-matched network that has never seen collider data.
  
\subsubsection{Neural Forecasting}

In the neural forecasting task, a model is trained on neural activity recorded from a cluster of neurons and is used to predict the future signals of the same cluster. The time-series data exhibit characteristic temporal patterns and pronounced correlations across channels which are hypothesized to be captured by the feature mapping in Table~\ref{tab:mapping}, allowing for transfer learning from NEXUS. 
The coefficient of determination $R^2$ between the predicted and ground-truth future activity is used as a performance metric, averaged over channels and time steps,
\begin{equation}
R^2 = 1 - \frac{\sum_i (y_i - \hat{y}_i)^2}{\sum_i (y_i - \bar{y})^2},
\end{equation}
where $y_i$, $\hat{y}_i$, and $\bar{y}$ denote the ground-truth, predicted, and mean activity, respectively. Evaluating the chained Stage-1 and Stage-2 models (see Section~\ref{sec:methods}) end to end, the design attains $R^2 = 0.891$ for subject~1 (\textit{affi}) and $R^2 = 0.892$ for subject~2 (\textit{beignet}). 
These results indicate strong predictive performance, demonstrating another successful mapping of out-of-domain data of comparable modality and complexity onto the collider-based pre-training of NEXUS.

\section{Discussion}
\label{sec:discussion}

Together the in- and out-of-domain results serve as strong support for NEXUS's capability as a foundation model. 
Evidence for successful transfer learning appears in the favorable data scaling
observed for every downstream task for which a training dataset scan is performed
(Figures~\ref{fig:FMplots_ATLAS} and~\ref{fig:FMplots_OOD}), where NEXUS achieves higher performance with fewer labeled examples.
This demonstrates that the pre-trained weights meaningfully contribute to tasks not directly related to the backbone training objective, i.e., the reconstruction of track features. 
Latent space studies indicate that the success of NEXUS can be attributed to the backbone learning key correlations between tracks that encode the detector response of SM and BSM processes, which not only enables event classification but also prediction of jet features despite the lack of calorimeter information.
The extent of NEXUS transfer learning to out-of-domain tasks in other physical systems can be understood by the design of the collider track-based pre-training to learn general, symmetry-aware structure in sparse, noisy, multi-scale signals.

Unique to this work is the fact that these results are achieved with an autoencoder of $\mathcal{O}(1)$ million parameters, offering a foundation model pathway with substantially lower computational overhead than its transformer-based peers.
To quantify this, a parameter-matched transformer baseline is trained on the same
20 million event pre-training sample used for NEXUS. The baseline comprises four
pre-layer-normalization self-attention blocks with $d_{\text{model}}=256$ (3.19 million encoder parameters) and the same 128-dimensional latent space.
Since every attention and feed-forward layer executes once per track token, the transformer costs 96.9 million multiply–accumulate operations (MMACs) (0.19 GFLOPs) per event at inference, against 3.1 MMACs (0.006 GFLOPs) for NEXUS; its unsupervised pre-training likewise required $\sim$17 GPU-minutes per epoch on a single NVIDIA A100, compared with $\sim$22 GPU-seconds per epoch for NEXUS ($\sim$46$\times$ faster), despite bf16 mixed precision for the transformer.
This efficiency does not come at the cost of in-domain transfer quality: frozen-encoder linear probes on 20-class SM classification perform comparably for the two backbones (within $\sim$2\%, with a slight edge to the transformer).

The DNN approach of NEXUS also opens the door to field-programmable gate array (FPGA) implementation with low latency, whereas low-latency transformer deployment remains impractical on standard FPGA devices, bringing the benefits of transfer learning to scientific data acquisition and filtering. 
This model design choice also mitigates data center power usage and financial cost for processing domain-specific data, which represent a significant challenge for the future of collider physics~\cite{Albrecht2019}.
Beyond deployment, the comparative simplicity of the NEXUS architecture translates into faster training and inference, lowering the barrier to large-scale development for scientific applications with limited computational access. This may further open new directions for foundation models that combine multiple architectures to accommodate different modalities and information densities, while balancing computational efficiency with practical performance.

Future work is needed to articulate the exact data regime in which autoencoder-based backbones are attractive. 
For track features, which are effectively 1D and sparse, clear benefits are observed from training a relatively simple architecture, but higher-dimensional, denser modalities (e.g., 2D images or dense 3D point clouds) may require additional architectural complexity to be fully characterized.
FPGA implementation would require model compression techniques such as pruning, quantization, or distillation to preserve performance within the resources of state-of-the-art devices~\cite{Deiana_2022}, along with co-design between algorithmic objectives and system constraints.

As one of the most data-intensive and potentially discovery-rich scientific disciplines, the use of foundation models in collider physics could substantially improve and simplify all experimental workflows.
Moreover, the transfer of track-based pre-trained knowledge to out-of-domain waveform tasks highlights the breadth of these learned representations and suggests a promising path for leveraging particle physics priors to benefit a wider range of scientific applications.
This offers a window into a future of scientific development with dense interdisciplinary collaboration leveraged for unprecedented access to new fundamental processes and technologies. 
\section{Methods}
\label{sec:methods}

\subsection{Dataset}
\label{subsec:dataset}

All samples are taken from ATLAS Open Data~\cite{atlas_opendata_website} at $\sqrt{s}=13$~TeV in the \texttt{DAOD\_PHYSLITE} format, comprising both recorded proton–proton collision events and MC-simulated processes. 
NEXUS is pre-trained on unlabeled collision data from the 2016 LHC Run~2 period, using the first eight of the 186 available run periods, from which approximately 20 million events are obtained.
The downstream classification studies use SM MC samples spanning 20 physics process categories, including $W/Z$+jets resolved by decay mode, diboson production ($WW$, $WZ$, $ZZ$), top-quark pair production ($t\bar{t}$), and quantum chromodynamics (QCD) multijet events. 
The BSM samples used for the anomaly detection and classification tasks comprise eight signal classes: heavy resonances with masses of 3~TeV ($Z'\to t\bar{t}$, $Z'\to b\bar{b}$, $Z'\to\tau^{+}\tau^{-}$, $Z'\to e^{+}e^{-}$, $W'\to q\bar{q}$, and $W'\to tb$, the latter treated as two separate classes according to whether the top quark decays hadronically or leptonically), together with 1~TeV leptoquarks. All signals are generated with a combination of \texttt{MadGraph 5}~\cite{Alwall_2011} and \texttt{Pythia 8}~\cite{SJOSTRAND2020106910}.

All studies use exclusively the \texttt{InDetTrackParticles} collection, from which six per-track features are extracted: the pseudorapidity $\eta=-\ln[\tan(\theta/2)]$ and transverse momentum \pt~$=|1/(q/p)|\sin\theta$, both derived from the fitted track parameters, together with the azimuthal angle $\phi$, the transverse and longitudinal impact parameters $d_0$ and $z_0$, and the track fit quality $\chi^2$. 
Each event passes through a uniform pipeline: tracks are required to satisfy $|d_0|<10$~mm to suppress poorly reconstructed
trajectories. They are then sorted by descending \pt, and truncated to the 30 highest-\pt~tracks; events with fewer than 30 tracks are
zero-padded and accompanied by Boolean masks so that padded entries are excluded throughout training. Each feature is then mapped to a comparable dynamic range by a fixed analytic transform tuned to the ATLAS tracker acceptance: $\eta$, $\phi$, $d_0$, $z_0$ and $\chi^2$ are linearly rescaled, while \pt~is logarithmically compressed to tame its heavy tail. No further dataset-level normalization is applied.

\subsection{Backbone Model}

NEXUS is a fully connected autoencoder trained to reconstruct the per-event track tensor. 
The $30\times6$ preprocessed tracks are flattened into a 180-dimensional input and progressively compressed by an encoder of three hidden layers of width 2048, 1024 and 512 into a
128-dimensional latent representation; an asymmetric decoder with hidden widths 128, 256 and 512 maps the latent back to the 180-dimensional reconstruction. All hidden layers use GELU activations, with progressive dropout ($0$, $0.05$, $0.1$ across encoder depth) and no batch normalization. The full autoencoder contains approximately 3.3 million parameters, of which the encoder ($\sim$3.1 million) constitutes the transferable backbone reused in every downstream task.

Reconstruction is optimized with a Huber loss evaluated element-wise and masked by the per-track Boolean masks, so that zero-padded tracks contribute neither to the loss nor to its gradients. Training uses the Adam optimizer~\cite{kingma2017adammethodstochasticoptimization} at a base learning rate of $8\times10^{-4}$ and a large batch size of $92{,}160$ events for up to $1{,}000$ epochs, on the pre-training set split $84/8/8\%$ into training, validation and test partitions with a fixed random seed. The learning rate is halved on validation plateaus (patience 10 epochs, floor $10^{-7}$) and training is terminated by early stopping with a patience of 30 epochs. The parameter-matched transformer baseline of Section~\ref{sec:discussion} is pre-trained on the identical sample, split and masked reconstruction objective, but configured for its own best training throughput: bf16 mixed precision, a batch size of $4{,}096$ events and an early-stopping patience of 15 epochs. Both models are trained on a single NVIDIA A100, and wall-clock cost is quoted per epoch, i.e., per complete pass over the pre-training partition, which is the unit independent of batch size.

\subsection{Downstream Tasks}

All downstream tasks share a common transfer protocol: the input is cast into the $30\times6$ track-tensor representation used in pre-training---directly for
collider data, and via the waveform descriptors of Table~\ref{tab:mapping} for the out-of-domain time series---encoded by the NEXUS backbone into the
128-dimensional latent space, and mapped to task outputs by a lightweight head.
Unless stated otherwise, each task is evaluated under the training schemes of
Section~\ref{sec:results} on identical data splits and as a function of the
number of labeled training events, so that performance differences isolate the
contribution of the pre-trained representation. In the \textit{fine-tuned} scheme the encoder and head are updated jointly with
discriminative learning rates, $1\times10^{-5}$ for the pre-trained encoder and
$1\times10^{-3}$ for the randomly initialized head, so that the pre-trained
features are refined rather than overwritten. The \textit{fixed backbone} and
\textit{randomly initialized} schemes train only the head, at $1\times10^{-3}$.
The \textit{from scratch} scheme trains the full network from random weights at
$1\times10^{-3}$. The remainder of this section
specifies the task-specific data, targets, and heads.

\subsubsection{Collider Physics} 
For the two collider domain classification tasks, signal classes can differ substantially in production rates and available statistics. 
To mitigate class-imbalance effects, each class is balanced to a common number of training events, with mild oversampling applied only to the rarest signals. Equal-size, held-out validation and test sets are constructed per class for all reported metrics, ensuring that accuracy reflects genuine class separability rather than relative class abundance.

In the jet regression task, all three targets refer to the leading (highest-\pt) jet of the event and are predicted jointly from a single NEXUS latent vector by two lightweight, task-specific heads sharing the same backbone: a regression head for the continuous jet kinematics (\pt~and $m$) and a three-way classification head separating non-$b$-jets, $b$-jets, and the no-jet case in which the event contains no reconstructed jet. For events with no reconstructed jet, only the classification head contributes and the regression loss is masked. 
Truth-level jet labels are drawn from a mixture of Standard Model processes spanning a broad kinematic and flavor range (QCD multijet, $W/Z$+jets, and $t\bar{t}$ in its fully hadronic, semi-leptonic and di-leptonic channels), so that both light- and heavy-flavor jets are well represented. The regression targets are learned in the same log-compressed space used for the track inputs and are reported in physical units (GeV) after inverting the transformation, while the two heads are trained jointly under a combined mean-squared-error and cross-entropy objective.

\subsubsection{Out-of-Domain} 
Gravitational waves, ripples in spacetime produced by the acceleration of massive objects, are observed by Laser Interferometer Gravitational-Wave Observatory (LIGO)~\cite{Abbott_2023} through the tiny strain $h(t)$ they induce across two widely separated interferometers. Well-modeled transients such as BBH mergers are recovered by matched filtering against banks of analytically or numerically computed templates, a strategy that is by construction insensitive to unmodeled or unforeseen sources for which no template exists. The National Science Foundation (NSF) Harnessing the Data Revolution (HDR) Anomaly Detection Challenge~\cite{nsfhdr_ad} targets precisely this regime, using whitened, band-pass-filtered strain from LIGO's third observing run (O3)~\cite{Abbott_2023} together with simulated signals injected into the real instrument noise.

The coastal flooding task considers anomalous sea-level variability, which arises from a complex interplay of large-scale atmospheric and ocean dynamics with local meteorological and coastal conditions, making it difficult to forecast with traditional statistical methods.
Decades of high-resolution sea-level time series data at coastal stations along the East Coast are provided with the goal of predicting coastal flooding events across spatially distinct locations~\cite{nsfhdr_mood}.
For each station, a seven-day (168-hour) window of hourly sea level is divided into 30 segments and summarized by the waveform descriptors of Table~\ref{tab:mapping}, forming the $30\times6$ track-like tensor expected by the backbone. A lightweight head maps the resulting NEXUS latent to the number of flooding days in the following 14-day window, predicted as a binned regression over the integer counts $0$ to $14$. 
Training and validation use a stratified temporal split of nine East-Coast stations, while the test set combines a held-out temporal segment of those stations with three further stations excluded from training entirely, so that evaluation probes both temporal and spatial generalization.

In neural forecasting, predicting the next frame of multi-channel neural activity from the preceding frames has emerged as an effective approach to modeling neural systems~\cite{le2022stndt,li2023amag,pandarinath2018inferring}. 
The dataset comprises neural activity recordings from two monkey subjects, \textit{affi} (239 channels) and \textit{beignet} (89 channels)~\cite{nsfhdr_mood}. 
Each instance consists of $10$ consecutive time steps of neural activity as input and the subsequent $10$ time steps as the forecasting target. 
The time-series data are mapped to track-segment-like features, as described in Table~\ref{tab:mapping}: each channel is treated as a single ``track'', and a sliding window groups $30$ neighboring channels into one training instance, ordered by descending signal variability. 
A two-stage model is adopted in which both stages use the pre-trained NEXUS model as the encoder and are fine-tuned on their respective objectives. 
The Stage-1 model predicts the future features of the same $30$ channels from their past features, while the Stage-2 model maps features to the raw time-series signal and is trained against the ground-truth future recordings. Chaining the two stages yields the complete forecasting pipeline while sharing the same time-series feature set.



\backmatter





\section*{Data \& Code Availability}

All data used in this study are publicly accessible~\cite{atlas_opendata_website, nsfhdr_ad, nsfhdr_mood}. 
All analysis code is publicly available at \url{https://github.com/SLAC-Julia-Group/NEXUS_public}.

\section*{Declarations}

\bmhead{Funding}

This work is supported by the U.S. Department of Energy under contract number DE-AC02-76SF00515 and the Office of the Vice Provost for Undergraduate Education at Stanford University.

\bmhead{Author contributions}

LW was the primary analyzer, developed the backbone and downstream models, prepared all final results and figures excluding the neural forecasting task, and wrote the manuscript. QL provided expertise on backbone design and optimization, aided in framework development, prepared the neural forecasting result, and contributed to paper writing. AY designed the sample processing framework, performed initial input studies, and contributed to collider downstream tasks. JG conceptualized the study, supervised the team, and led manuscript preparation. 

\bmhead{Competing interests}

The authors declare no competing interests.

\bibliography{main}

\end{document}